\documentclass[journal]{IEEEtran}

\usepackage{physics}
\usepackage{graphicx}
\usepackage{float}
\usepackage{amsmath}
\usepackage{times}
\usepackage{subfigure}
\usepackage{amsmath}
\usepackage{longtable}
\usepackage{booktabs}
\usepackage{float}
\usepackage{threeparttable}
\usepackage{xcolor}
\usepackage{diagbox}
\usepackage{cite}
\usepackage{color}
\usepackage{tabularx}
\usepackage{multirow}
\usepackage{mathpazo}
\usepackage{mathrsfs}

\usepackage{amsfonts}
\usepackage{amsmath}
\usepackage{amssymb}
\usepackage{makecell}
\usepackage{xspace}
\usepackage[breaklinks=true,colorlinks,bookmarks=false]{hyperref}

\makeatletter
\DeclareRobustCommand\onedot{\futurelet\@let@token\@onedot}
\def\@onedot{\ifx\@let@token.\else.\null\fi\xspace}
\def\eg{\emph{e.g}\onedot} 
\def\ie{\emph{i.e}\onedot}

\makeatother

\ifCLASSINFOpdf

\else

\fi

\begin{document}
\pagestyle{empty}

\title{VehicleGAN: Pair-flexible Pose Guided Image Synthesis for Vehicle Re-identification}  

\author{
Baolu Li$^{\dagger}$,  
Ping Liu$^{\dagger}$,
Lan Fu, 
Jinlong Li, 
Jianwu Fang, 
Zhigang Xu$^{*}$, 
Hongkai Yu$^{*}$% <-this % stops a space
\thanks{Baolu Li and Zhigang Xu are with Chang'an University, Xi’an 710064, China. Ping Liu is with the Center for Frontier AI Research (CFAR), Agency for Science, Technology, and Research (A*STAR), Singapore 138634. Lan Fu is with University of South Carolina, Columbia 29201, SC, USA. Jianwu Fang is with Xi'an Jiaotong University, Xi'an 710049, China. Baolu Li, Jinlong Li, and Hongkai Yu are with Cleveland State University, Cleveland, OH 44115, USA.}
\thanks{${\dagger}$ indicates co-first authors. * Co-corresponding authors: Zhigang Xu (xuzhigang@chd.edu.cn), Hongkai Yu (h.yu19@csuohio.edu).} 
% \thanks{Fundings TBD.}
}% <-this % stops a space


\maketitle 
\thispagestyle{empty}
\begin{abstract}
Vehicle Re-identification (Re-ID) has been broadly studied in the last decade; however, the different camera view angles leading to confused discrimination in the feature subspace for the vehicles of various poses, is still challenging for the Vehicle Re-ID models in the real world. To promote the Vehicle Re-ID models, this paper proposes to synthesize a large number of vehicle images in the target pose, whose idea is to project the vehicles of diverse poses into the unified target pose so as to enhance feature discrimination. Considering that the paired data of the same vehicles in different traffic surveillance cameras might be not available in the real world, we propose the first Pair-flexible Pose Guided Image Synthesis method for Vehicle Re-ID, named as VehicleGAN in this paper, which works for both supervised and unsupervised settings without the knowledge of geometric 3D models. Because of the feature distribution difference between real and synthetic data, simply training a traditional metric learning based Re-ID model with data-level fusion (\ie, data augmentation) is not satisfactory, therefore we propose a new Joint Metric Learning (JML) via effective feature-level fusion from both real and synthetic data. Intensive experimental results on the public VeRi-776 and VehicleID datasets prove the accuracy and effectiveness of our proposed VehicleGAN and JML.          

\end{abstract}


\IEEEpeerreviewmaketitle

%%%%%%%%% BODY TEXT
%-----------------------------------------------------------------------
%%%%%%%%% Introduction 
\section{Introduction}

Many tasks and functions in the intelligent transportation systems~\cite{teng2023motion,10363115,10232886,10415232,10380440,10470374,10009795,10058739,9082629,10332939,10239114,10057448,10287578} involve the detection or identification of vehicles.  
Vehicle Re-identification (Re-ID) is an important task in intelligent transportation systems, as it allows for the retrieval of the same vehicle from multiple non-overlapping cameras. With the availability of vehicle surveillance datasets~\cite{liu2016deep1, liu2016deep2,lou2019veri}, many vehicle Re-ID methods~\cite{10173635,10374282,10382666} have been proposed, which have gained wide interest among the research communities of foundation intelligence, human-machine systems, and transportation~\cite{10138462,10058739,9684293}. 

\begin{figure}[!t]
\centering
\includegraphics[width=1\columnwidth]{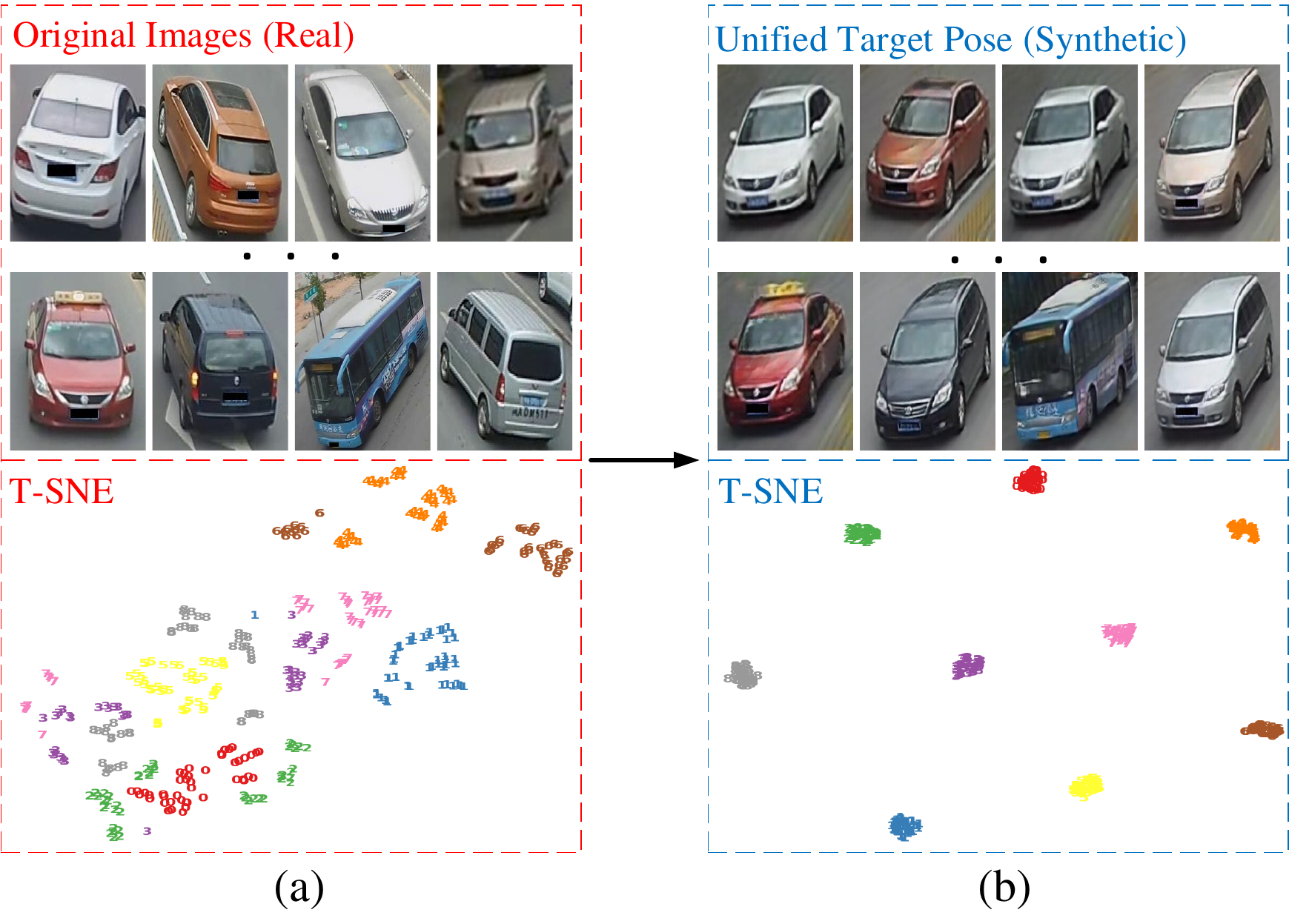}
\DeclareGraphicsExtensions.
\vspace{-16pt}
\caption{Enhanced discrimination by synthesizing vehicle images to a unified target pose by our VehicleGAN. (a) Vehicle images in the VeRi-776~\cite{liu2016deep1}. (b) Corresponding synthesized vehicle images by VehicleGAN.}\vspace{-16pt}
\label{fig:introduction_1}
\end{figure}

However, the large viewpoint divergence of vehicle images caused by different camera views in the real world makes significant challenges for vehicle Re-ID models~\cite{liu2016deep,liu2016deep1}. As shown in Fig.~\ref{fig:introduction_1}, the same vehicles of diverse poses are ambiguous in an embedded feature subspace, leading to identification difficulties, while the feature discrimination could be enhanced if the vehicle images could be  projected to the same target pose. Inspired by our discovery of Fig.~\ref{fig:introduction_1}, this paper proposes to project the vehicles of diverse poses into the unified target pose.

To tackle the pose-varied vehicle images effectively, the controllable various-view synthesis of vehicle images has been investigated~\cite{lv2020pose}, which aims to synthesize the images of a vehicle at a target pose. Existing methods can be divided into 3D-based and 2D-based approaches. Those 3D-based approaches~\cite{rematas2016novel} utilize the geometric 3D model to synthesize images, which might be not available or prone to errors in the real traffic surveillance scenarios due to the lack of camera parameters and diverse vehicle poses. The 2D-based methods~\cite{lv2020pose} use paired 2D  images of the same vehicle in different cameras to supervise neural  networks to learn the transformation of the vehicle to the target pose. They suffered from the manual annotation cost of the same vehicles in different cameras. Therefore, no matter the existing 3D or 2D methods have significant drawbacks in the real world.

Differently, this paper proposes the first Pair-flexible Pose Guided Image Synthesis method for Vehicle Re-ID, named as VehicleGAN, which works for both supervised and unsupervised settings without the knowledge of geometric 3D models. 1) We design a novel Generative Adversarial Network (GAN) based end-to-end framework for pose guided vehicle image synthesis, which takes the 2D vehicle image in the original pose and the target 2D pose as inputs and then directly outputs the new synthesized 2D vehicle image in the target pose. Using the 2D target pose as a condition to control the Generative Artificial Intelligence (AI), the proposed method gets rid of using the geometric 3D model. 2) The proposed VehicleGAN works for both supervised (paired images of same vehicle) and unsupervised (unpaired images of same vehicle) settings, so it is called Pair-flexible in this paper. For the pose guided image synthesis in the current vehicle Re-ID research community, the supervised (paired) setting is easy for training the Generative AI model, however, the unsupervised (unpaired) setting is challenging. 3) To solve the challenging unsupervised problem, we proposed a novel method AutoReconstruction to transfer the vehicle image in the original pose to the target pose and then transfer it back to reconstruct itself as self-supervision. In this way, the paired images of the same vehicles in different cameras are not required to train the Generative AI model.

After getting the synthesized vehicle images in different poses, simply training a traditional metric learning based Re-ID model with the direct data-level fusion of real and synthetic images (\ie, data augmentation) is not satisfactory. This is because of the feature distribution difference between real and synthetic data. To solve this problem, we propose a novel Joint Metric Learning (JML) via effective feature-level fusion from both real and synthetic data. We conduct intensive experiments on the public VeRi-776~\cite{liu2016deep1} and VehicleID~\cite{liu2016deep2} datasets, whose results display the accuracy and effectiveness of our proposed VehicleGAN and JML. The main contributions are summarized as follows.

\begin{itemize}
\item This paper proposes a novel method to project the vehicles of diverse poses into the unified target pose to enhance vehicle Re-ID accuracy.

\item This paper proposes the \textbf{first} Pair-flexible Pose Guided Image Synthesis method for Vehicle Re-ID, called VehicleGAN, which works for both supervised and unsupervised settings without the knowledge of geometric 3D models.

\item This paper proposes a new Joint Metric Learning (JML) via effective feature-level fusion from both real and synthetic data to overcome the shortcomings of the real-and-synthetic feature distribution difference.   

\end{itemize}

\section{Related Works}

\subsection{Vehicle Re-ID}

Benefiting from a series of public datasets and benchmarks~\cite{liu2016deep1,liu2016deep2}, vehicle Re-ID has made significant progress over the past decade. All the previous works of vehicle Re-ID task are to enhance the feature discrimination of vehicles in different cameras. On the one hand, some previous works~\cite{wang2017orientation} aim to learn supplemental features of the local vehicle regions to reinforce the global features. On the other hand, some previous works~\cite{chen2019partition} focus on designing more powerful neural network structures to enhance feature discrimination. Different with existing feature enhancement works, this paper proposes to project the vehicles of diverse poses into the unified pose so as to enhance feature discrimination for vehicle Re-ID.

\subsection{Pose Guided Vehicle Image Synthesis}
Pose Guided Vehicle Image Synthesis allows vehicles to synthesize novel views based on pose. Previous works can be mainly divided into 3D-based and 2D-based approaches. Those 3D-based methods~\cite{rematas2016novel}rely on 3D model of vehicle or parameters of camera to achieve perspective conversion. 3D-based methods are limited by the difficulty to obtain detailed 3D models or accurate camera parameters in real scenes. Benefited from the Generative AI, the 2D-based methods~\cite{lv2020pose} learn the experience of view synthesis from paired 2D images in various poses. These 2D-based methods can more easily extract poses from pictures, which will be more advantageous in the real world. However, since they require ground-truth to supervise the learning, the paired images of same vehicle need to be manually identified. Differently, our proposed method is less constrained in the real world and can be unsupervised without the need for identity annotations.

\begin{figure}[!t]
\centering
\includegraphics[width=1\columnwidth]{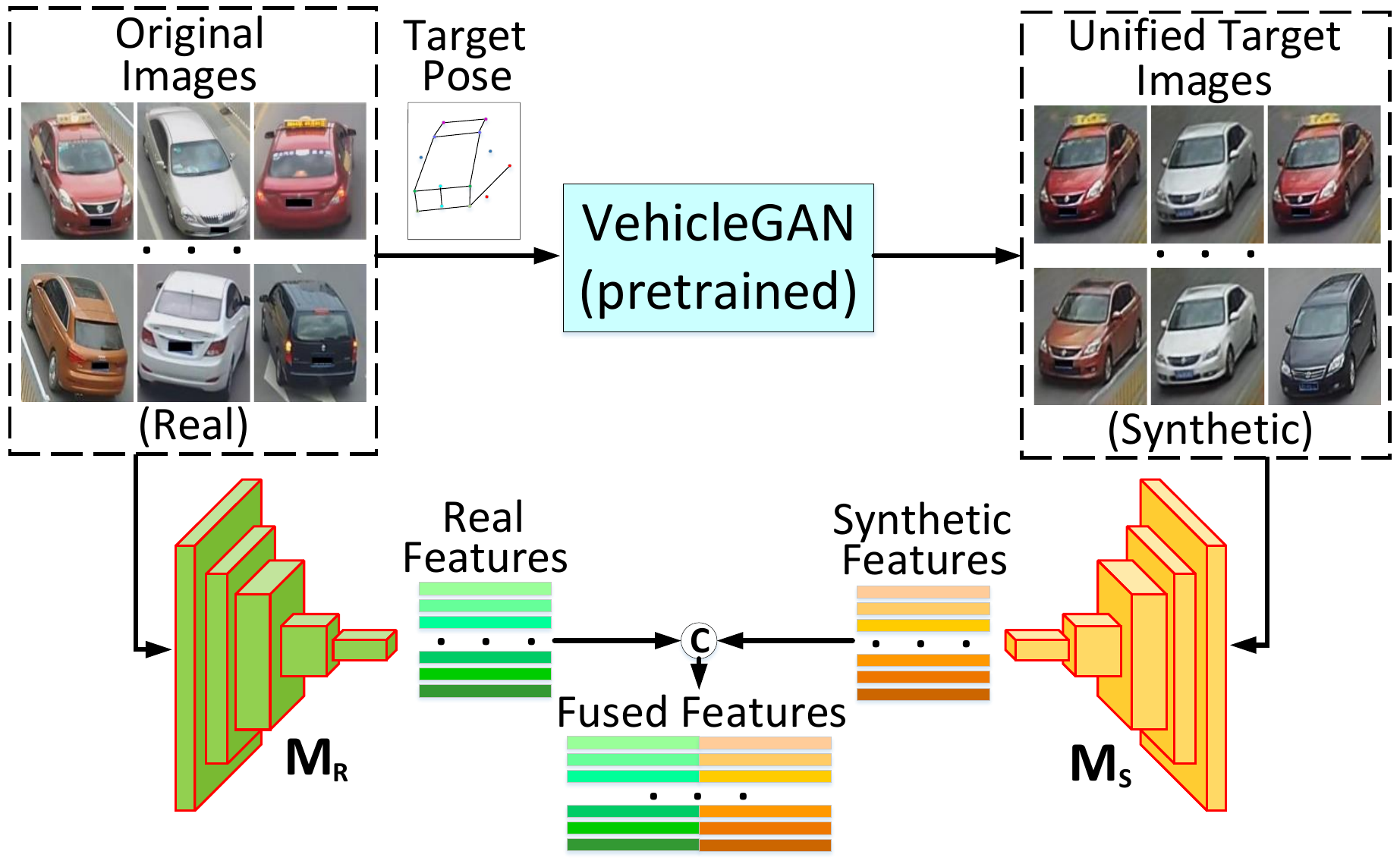}
% \DeclareGraphicsExtensions.
\caption{\textbf{The pipeline of the proposed Joint Metric Learning using the proposed VehicleGAN.} $\mathrm{M_R}$: Re-ID  model for real images. $\mathrm{M_S}$: Re-ID  model for synthetic images. The Unified Pose part represents the synthetic images with the same target pose by the proposed VehicleGAN (pre-trained). The real and synthetic features are fused for joint training and testing.}\vspace{-16pt}
\label{fig:frame}
\end{figure}

\begin{figure*}[!t]
\centering
\includegraphics[width=0.9\textwidth]{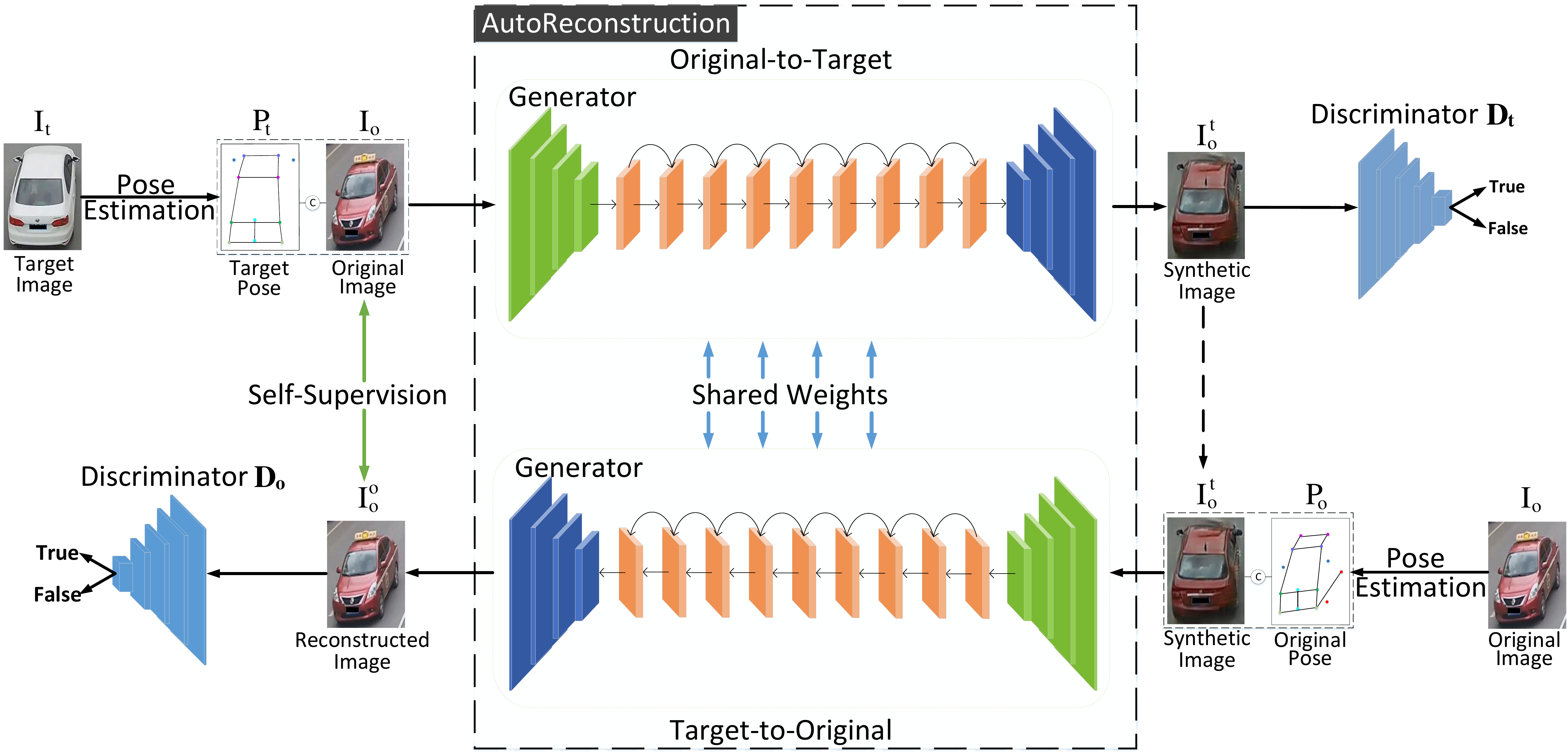}
\DeclareGraphicsExtensions.
\caption{\textbf{The detailed pipeline of the proposed VehicleGAN with the idea of AutoReconstruction.} The input of the generator is the channel-wise concatenation of the original image and the target pose, and the output of the generator is the synthesized image of the original image in the target pose. Unsupervised setting (unpaired data) is shown in this example.}\vspace{-16pt}
\label{fig:vehicleGAN}
\end{figure*}

\section{Proposed Approach}

\subsection{Overview}\label{sec:overview} 
The Fig.~\ref{fig:frame} shows the whole framework of the proposed VehicleGAN  guided vehicle Re-ID via Joint Metric Learning. The framework includes two stages: VehicleGAN and Joint Metric Learning. The former is an encoder-decoder based Generative Adversarial Networks (GAN) for pose guided vehicle image synthesis, and the latter consists of two branches: a Re-ID  model for real images ($\mathrm{M_R}$) and the other Re-ID  model for synthetic images ($\mathrm{M_S}$). The VehicleGAN aims to generate an image with a target pose given an original vehicle image as input. The two Re-ID  models (\ie, $\mathrm{M_R}$ and $\mathrm{M_S}$) do not share weights due to their feature differences. With the synthetic vehicle images in the unified target pose generated by VehicleGAN, the $\mathrm{M_S}$ learns to identify pose-invariant features, while the $\mathrm{M_R}$ learns to recognize features from real images. The $\mathrm{M_R}$ and $\mathrm{M_S}$ are trained by Joint Metric Learning.
The training of the whole pipeline of VehicleGAN can be implemented in an unsupervised way or a supervised way, which is named as Pair-flexible here.

\subsection{VehicleGAN}\label{sec:vehicleGAN}
Our VehicleGAN aims to generate synthetic images of a same vehicle under a specific target pose. As shown in Fig.~\ref{fig:vehicleGAN}, the VehicleGAN includes one generator and two discriminators. Given an original image $I_o$ and an image $I_t$ with the target pose, we first extract the target pose $P_t$ via a pose estimator $\psi$, \ie, $P_t=\psi(I_t)$. Then, a synthetic image $I_o^t$ will be generated via the generator $G$, \ie, $I_o^t=G(I_o, P_t)=G(I_o, \psi(I_t))$. $I_o^t$ denotes an image which has the content of image $I_o$ as well as the target pose of image $I_t$. A discriminator $D_{t}$ is to distinguish the synthetic fake $I_o^t$ from the real $I_t$. By our proposed AutoReconstruction, we reuse the generator to transfer $I_o^t$ back to an image $I_o^o$ with the content and pose of the original image $I_o$. The other discriminator $D_{o}$ is to distinguish the synthetic fake $I_o^o$ from the real $I_o$. Because of the proposed AutoReconstruction, we can supervisedly or unsupervisedly train the whole VehicleGAN pipeline.

\subsubsection{Pose Estimation}
Here, we use 20 keypoints~\cite{wang2017orientation} annotated on the VeRi-776~\cite{liu2016deep1} dataset to represent the vehicle pose. These keypoints are some discriminative positions on the vehicle, \eg, wheels, lights, and license plates. Specifically, we adopt the Deconvolution Head Network~\cite{xiao2018simple} as $\psi$ to estimate the vehicle pose by outputting a response map for each of the 20 keypoints. The response maps have Gaussian-like responses around the locations of keypoints. Given a target image $I_t$, the output pose response map is $P_t=\psi(I_t)$ with 20 channels.

\subsubsection{AutoReconstruction as 
Self-Supervision}
Given an original vehicle image $I_o$, and a target vehicle image $I_t$, the generator aims to synthesize a vehicle image with $I_o$ content in the $I_t$ pose. The input of the generator is the concatenation of the original image and the target pose. The generator adopts an encoder-decoder based network like PN-GAN~\cite{qian2018pose}. Reversely, the synthesized $I_o^t$ will go through the generator to reconstruct the original image using the pose of $I_o$, which generates the reconstructed image $I_o^o$. The Original-to-Target and Target-to-Original bidirectional image transfer is named as Autoreconstruction in this paper. Because of the designed Autoreconstruction, the original $I_o$ and  reconstructed $I_o^o$ can be forced to be identical as self-supervision. 

\subsubsection{Pair-flexible Settings}
The optimization of our VehicleGAN can be performed in either supervised or unsupervised way. When the original image $I_{o}$ and the target  image $I_{t}$ are from the same vehicle, the corresponding ground truth image of the generated image $I_{o}^t$ will be $I_{t}$, which can provide full supervision information for training. Meanwhile, the original image $I_{o}$ and the target image $I_{t}$ can be from different vehicles (unpaired) for unsupervised learning. This advanced pair-flexible setting is excellent for real-world usages.

\subsubsection{Supervised Learning with Paired Data}
To optimize the whole pipeline, we adopt four loss functions: adversarial loss, pose loss, identity-preserving loss and reconstruction loss. Please note that the loss functions using the paired data of the same identity are denoted as $\mathfrak{L}$ (supervised), while other loss functions are denoted as $\mathcal{L}$ (unsupervised) in this  paper.

\textbf{Adversarial Loss:} The adversarial losses $\mathcal{L}_{adv_1}$ and $\mathcal{L}_{adv_2}$ aim to make the synthetic images more similar to the real images. In specific, we want to align the generated images $I_o^t$ with $I_t$ and $I_o^o$ with $I_o$ via $\mathcal{L}_{adv_1}$ and $\mathcal{L}_{adv_2}$, respectively. We adopt two discriminators to perform the distribution alignment to distinguish real or fake. The optimizing object function is
\begin{align}
    \mathcal{L}_{adv_1} &=  \mathbb{E}_{I_t\sim p_{data}(I_t)}[\mathrm{log}D_t(I_t)] \nonumber \\
    &+ \mathbb{E}_{I_o^t\sim p_{data}(I_o^t)}[\mathrm{log}(1-D_{t}(G(I_o,P_t)))], \\
    \mathcal{L}_{adv_2} &=  \mathbb{E}_{I_o\sim p_{data}(I_o)}[\mathrm{log}D_o(I_o)] \nonumber \\
    &+ \mathbb{E}_{I_o^o\sim p_{data}(I_o^o)}[\mathrm{log}(1-D_{o}(G(I_o^t,P_o)))].
\end{align}

\textbf{Pose Loss:} $\mathcal{L}_{pose}$ is to align the poses of the synthetic images ($I_o^t$, $I_o^o$) with the guided poses during the Autoreconstruction. The pose loss is defined as

\begin{align}
\small
\mathcal{L}_{pose}(\psi, I_o^t, P_t, I_o^o, P_o) = \|\psi(I_o^t)-P_t\|_{2} + \|\psi(I_o^o)-P_o\|_{2}.
\end{align}

\textbf{Identity-preserving  Loss:} During the image transfer process of the VehicleGAN for the target pose, the vehicle identity information should be preserved, \ie, keeping the identity of the synthetic image consistent with that of the original image, \eg, $I_o^t$ and $I_o$. After pose synthesis of vehicles, the semantic content, style, texture, and color of the synthetic image should be kept. Therefore, we introduce \textit{style loss},  \textit{perceptual loss}, \textit{content loss} to optimize the network to preserve the identity.

We introduce Gram matrix~\cite{gatys2015texture}, which generally represents the style of an image, to construct the style loss $\mathcal{L}_{style}$. 
Let $\phi_{j}(I)$ $\in H_{j}\times W_{j} \times C_j$ be the feature map at $j$-th layer of VGG network for the input image $I$, then the Gram matrix is defined as a $C_{j}\times C_{j}$ matrix whose elements are given by
%----------------
\begin{align}
\mathcal{G}_{j}(I)_{c,c^{'}}&=\frac{1}{C_{j}H_{j}W_{j}}\sum_{h=1}^{H_{j}}\sum_{w=1}^{W_{j}}(\phi_{j}(I)_{h,w,c}\cdot\phi_{j}(I)_{h,w,c^{'}}).
\end{align}
%-----------
Then, the \textit{style loss} is formulated as the mean squared error between the Gram matrices of $I_{o}^t$ and $I_{t}$ as 
%-----------
\begin{align}
\mathfrak{L}_{style}=\sum_{j}\|\mathcal{G}_{j}(I_{o}^t)-\mathcal{G}_{j}(I_{t})\|_2,
\end{align}
%------------
where we use the feature maps of [$relu$1\_1, $relu$2\_1, $relu$3\_1, $relu$4\_1] layers to calculate the style loss.

We define the \textit{perceptual loss} as
\begin{align}
\mathfrak{L}_{per}=\|\phi_{j}(I_{o}^t)-\phi_{j}(I_{t})\|_{2},
\end{align}
where we use the feature map from the $relu$4\_1 layer of VGG network to compute the perceptual loss. Also, the reconstructed image is expected to keep the same content as the source image, then, the \textit{content loss} is defined as
\begin{align}
\mathcal{L}_{c} = \sum_{j}^{}\|\phi_{j}(I_{o}^{o})-\phi_{j}(I_{o})\|_{2},
\end{align}
where we use the feature maps from [$relu$1\_1, $relu$2\_1, $relu$3\_1, $relu$4\_1] layers of VGG network. Therefore, the identity-preserving loss is formulated to the weighted sum of the above three losses as 
\begin{align}
\mathfrak{L}_{idp}=\beta _{1}\mathfrak{L}_{style}+\beta _{2}\mathfrak{L}_{per}+\beta _{3}\mathcal{L}_{c}.
\end{align}

\textbf{Reconstruction Loss:} We employ reconstruction loss to measure the pixel-wise error between the generated images and their ground truth, which is defined as  
\begin{align}
\mathfrak{L}_{rec}=\|I_{o}^{o}-I_{o}\|_{1}+\delta \|I_{o}^{t}-I_{t}\|_{1}.
\end{align}
On summary, we define the total supervised loss ${Loss}_{sp}$ as a weighted sum of all the defined losses:
\begin{align}
\footnotesize
{Loss}_{sp}=\lambda _{1}\mathcal{L}_{adv_{1}}+\lambda _{2}\mathcal{L}_{adv_{2}}+\lambda _{3}\mathcal{L}_{pose}+\lambda _{4}\mathfrak{L}_{idp}+\lambda _{5}\mathfrak{L}_{rec}.
\end{align}

\subsubsection{Unsupervised Learning with Unpaired Data} 
The input original image $I_{o}$ and the target image $I_{t}$ might be from different vehicle identities. The generated image $I_{o}^t$ does not have ground truth for supervision. In this way, the VehicleGAN can only be optimized in an unsupervised way. Since the style loss $\mathfrak{L}_{style}$, perceptual loss $\mathfrak{L}_{per}$, reconstruction loss $\mathfrak{L}_{rec}$ require the ground truth of $I_{o}^t$ for computation, we need to reformulate the three losses. In addition, because of the lack of supervision, we propose a trust-region learning method to reduce the degradation effects of the background region of different vehicles in image transfer.

\textbf{Trust-region Learning:} We propose a trust-region learning method to only focus on the trust regions (\ie, the shape of vehicle) in the unsupervised setting. We utilize 20 keypoints of a vehicle to  represent vehicle pose. We use the positions of these keypoints to calculate the convex hull surrounding the vehicle as mask. Let $M \in \mathbb{R}^{1\times H\times W}$ represents a binary mask formed by the pose $P$, where $H$ and $W$ represent the height and width of the pose feature maps. The values inside the convex hull/shape of the vehicle are all set as 1 (trust regions) while be 0 when outside of the convex hull. Then, $M$ is inferred from the size of feature maps of $P$ through average pooling to represent the size of the vehicle image.

\textbf{Losses Reformulation:} Due to the lack of paired data of same vehicles, we propose a trust-region style loss $\mathcal{L}_{style}$, a trust-region perceptual loss $\mathcal{L}_{per}$, a new reconstruction loss $\mathcal{L}_{rec}$  to replace the $\mathfrak{L}_{style}$, $\mathfrak{L}_{per}$,   $\mathfrak{L}_{rec}$ to optimize VehicleGAN in an unsupervised way.

Given the Gram matrix, we define a trust-region Gram matrix to calculate the style loss, which is defined as
 
\begin{equation}
\small 
\mathcal{G}_{j}(I,M)_{c,c^{'}}=\frac{1}{C_{j}H_{j}W_{j}}\sum_{h=1}^{H_{j}}\sum_{w=1}^{W_{j}}(\phi_{j}(I)_{h,w,c} \cdot M)\cdot(\phi_{j}(I)_{h,w,c^{'}} \cdot M). 
\end{equation} 

The proposed trust-region style loss is formulated to
\begin{align} 
\mathcal{L}_{style}=\sum_{j}\|\mathcal{G}_{j}(I_{o}^t, M_t)-\mathcal{G}_{j}(I_{o}, M_o)\|_2,
\end{align}
where $M_t$ and $M_o$ represents the trust-region masks corresponding to the pose of images $I_{o}^t$ and $I_{o}$, respectively.

The trust-region perceptual loss is formulated to
\begin{align}
\mathcal{L}_{per}=\|\phi_{j}(I_{o}^t)\cdot M_t-\phi_{j}(I_{o})\cdot M_o\|_{2}.
\end{align}

Then, we can replace the supervised loss $\mathfrak{L}_{idp}$ as $\mathcal{L}_{idp}=\beta _{1}\mathcal{L}_{style}+\beta _{2}\mathcal{L}_{per}+\beta _{3}\mathcal{L}_{c}$ in an unsupervised manner. The unsupervised reconstruction loss is re-defined as self-supervision only via  
\begin{align}
\mathcal{L}_{rec}=\|I_{o}^{o}-I_{o}\|_{1}.
\end{align}

The total unsupervised loss ${Loss}_{usp}$ is reformulated to 
\begin{align}
\footnotesize
{Loss}_{usp}&=\lambda _{1}\mathcal{L}_{adv_{1}}+\lambda _{2}\mathcal{L}_{adv_{2}}+\lambda _{3}\mathcal{L}_{pose}+\lambda _{4}\mathcal{L}_{idp}+\lambda _{5}\mathcal{L}_{rec}.  
\end{align}

\subsection{Joint Metric Learning}\label{sec:joint}

Given a pre-trained VehicleGAN obtained in Sec.~\ref{sec:vehicleGAN}, we first synthesize a unified target-pose image for each original vehicle image. Then, the original real images are fed into the Re-ID model $\mathrm{M_R}$, and the synthetic images with unified pose go through the Re-ID model  $\mathrm{M_S}$. $\mathrm{M_R}$ and $\mathrm{M_S}$ are optimized by a Joint Metric Learning (JML).

\subsubsection{Unified Target Pose}
Following~\cite{liu2016deep1}, we classify vehicles into nine categories, \ie, sedan, suv, van, hatchback, mpv, pickup, bus, truck, and estate. We manually choose one target-pose image for each of the nine categories as the unified target-pose image. Then, each original image can be translated into a synthetic image with the unified target pose by VehicleGAN, as shown in Fig.~\ref{fig:frame}.

\subsubsection{Re-ID Model}
We adopt ResNet50 as backbone for $\mathrm{M_R}$ and $\mathrm{M_S}$ models. Besides, we modify the stride of the last convolutional layer of the network to 1 to obtain larger-size feature maps with rich information. For the whole pipeline, the input vehicle image goes through the model  to obtain a 2048-dimensional feature map with size $16\times16$. Then, the feature map goes through a global average pooling layer to output a 2048-dimensional feature vector $f$. Thus, the original image and the synthetic image with the unified pose are fed into $\mathrm{M_R}$ and $\mathrm{M_S}$ to obtain feature vectors $f_{r}$ and $f_{s}$, respectively. Then, the $f_{r}$ and $f_{s}$ are concatenated into a 4096-dimensional feature vector $f_{c}$ as the fused feature.

\subsubsection{Loss Functions}
We adopt two kinds of loss functions: triplet loss for metric learning and cross-entropy loss for classification. The total loss function of JML is the sum of the four losses as follow: 

\begin{align}
{Loss}_{JML}={L}_{t_r}+{L}_{id_r}+{L}_{t_c}+{L}_{id_c}, 
\end{align}
where ${L}_{t_r}$ is triplet loss for real image features, ${L}_{id_r}$ is cross-entropy loss for real image features, ${L}_{t_c}$ is triplet loss for combined (real and synthetic) image features, ${L}_{id_c}$ is cross-entropy loss for combined image features.

\section{Experiments}

\subsection{Datasets and Evaluations}
We perform pose guided vehicle image synthesis and vehicle Re-ID on two public benchmark datasets VeRi-776~\cite{liu2016deep1} and VehicleID~\cite{liu2016deep2} for performance evaluation.

\subsubsection{VehicleGAN Experiments Setting} For the pose guided vehicle image synthesis task,  paired images from the same vehicle are inputs to the VehicleGAN for supervised learning, \textit{i.e.}, one is as the input original image, and the other is as the target pose image, which is also the ground truth for the synthesized image. For unsupervised learning, unpaired images from the same-type vehicle are fed into the VehicleGAN optimized through the unsupervised losses. During the inference stage, paired images from the same vehicle are fed into VehicleGAN for view synthesis and performance evaluation, \ie, the target pose image is the ground truth for the synthetic image after view synthesis.

\subsubsection{Datasets} 
\textit{VeRi-776}~\cite{liu2016deep1} includes more than 50,000 images of 776 vehicles. Following~\cite{liu2016deep}, VeRi-776 is split into a training subset (37,778 images of 576 vehicles) and a testing subset. \textit{VehicleID}~\cite{liu2016deep2} has 211,763 images with 26,267 vehicles. There are three test subsets with different sizes, \ie, \textit{Test800}, \textit{Test1600}, and \textit{Test2400}, for evaluation.

\subsubsection{Evaluation Metrics}
The evaluation metrics of pose guided vehicle image synthesis quality include Structural Similarity (SSIM) and Frechet Inception Distance (FID). The evaluation metrics of vehicle Re-ID accuracy  include mean Average Precision (mAP) and Cumulative Matching Characteristic (CMC) at Rank-1 and Rank-5.

\subsection{Implementation Details}

\subsubsection{VehicleGAN}
The resolution of input image in the VehicleGAN is $256\times 256$. For supervised learning, we set the loss weight parameters $\lambda _{1}$, $\lambda _{2}$, $\lambda _{3}$, $\lambda _{4}$, and $\lambda _{5}$ to 1, 0.2, 10,000, 1, 2, respectively. $\beta _{1}$, $\beta _{2}$, and $\beta _{3}$ are 1,000, 0.5, 0.05, respectively. $\delta$ is set to 4. For unsupervised learning, $\lambda _{1}$, $\lambda _{2}$, $\lambda _{3}$, $\lambda _{4}$, $\lambda _{5}$, $\beta _{1}$, $\beta _{2}$, and $\beta _{3}$ are 5, 1, 20,000, 1, 0.5, 500, 0.01, 0.1, respectively. We adopt Adam as the optimizer, and set the batch size to 12. We trained 200K iterations for supervised learning, and 300K  iterations for unsupervised learning.

\subsubsection{Re-ID model}
We utilize ResNet50 as the backbone network for $\mathrm{M_R}$ and $\mathrm{M_S}$, which is per-trained on ImageNet. The input image is resized to $224\times224$ before fed into the model. We set the batch size to 64, which includes 16 vehicle IDs, and 4 vehicle images for each vehicle ID. We perform data augmentation with random horizontal flipping, random cropping, and random erasing during training. We trained the $\mathrm{M_R}$ for 80 epochs when only the original image is fed into $\mathrm{M_R}$, and 100 epochs when the original image and synthetic image are fed into $\mathrm{M_R}$ and $\mathrm{M_S}$, respectively.

\subsection{Comparison for Pose Guided Vehicle Image Synthesis}
For supervised learning, we compare the proposed method with SOTA (state of the art) methods CGAN~\cite{mirza2014conditional}, PG2~\cite{ma2017pose}, DSC~\cite{siarohin2018deformable}, and PAGM~\cite{lv2020pose}. For unsupervised learning, we compare the proposed method with Perspective Transformation (PerTransf)~\cite{castleman1996digital}. We calculate the SSIM and FID metrics between the synthetic image and the target pose image, \ie, ground truth, for performance evaluation. As shown in Table~\ref{tab:view_syn}, our method gain the best performance in both supervised and unsupervised setting on SSIM and FID metric.

\begin{table}[ht]
\caption{Quantitative results on VeRi-776 and VehicleID. VehicleGAN and VehicleGAN* represent training in supervised and unsupervised manners, respectively.}
\resizebox{0.95\columnwidth}{!}{ 
\begin{tabular}{c|c|cc|cc} 
\toprule
\centering
\multirow{2}{*}{Strategies} & \multirow{ 2}{*}{Methods} & \multicolumn{2}{c|}{VeRi-776~\cite{liu2016deep1}} & \multicolumn{2}{c}{VehicleID~\cite{liu2016deep2}}  \\
& & SSIM $\uparrow$ & FID $\downarrow$ & SSIM $\uparrow$ & FID $\downarrow$ \\
\midrule
\multirow{ 5}{*}{Supervised}    &
CGAN\cite{mirza2014conditional} & 0.468 & 339.2 & 0.447 & 325.5 \\
&PG2\cite{ma2017pose} & 0.465 & 335.7 & 0.426 & 330.4 \\
&DSC\cite{siarohin2018deformable} & 0.456 & 305.7 & 0.425 & 320.7 \\
&PAGM\cite{lv2020pose} & 0.492 & 245.3 & 0.444 & 310.5 \\
&VehicleGAN & \textbf{0.554} & \textbf{233.0} & \textbf{0.551} & \textbf{193.6} \\
\midrule
\multirow{ 2}{*}{Unsupervised}  
& PerTransf~\cite{castleman1996digital} & 0.059 & 598.4 & 0.048 & 521.6 \\
& VehicleGAN* & \textbf{0.437} & \textbf{285.0} & \textbf{0.430} & \textbf{238.9} \\
\bottomrule
\end{tabular}}
\label{tab:view_syn}
\end{table}

\subsection{Comparison for Vehicle Re-ID}
The $\mathrm{M_R}$ model, optimized when only the original images are fed into the model, is the \textit{Baseline-ResNet50} method. When the original images and synthetic images are inputs for $\mathrm{M_R}$ and $\mathrm{M_S}$, respectively, \ie, $\mathrm{M_R}$ and $\mathrm{M_S}$ models are optimized together by involving pretrained VehicleGAN and Joint Metric Learning, the method is denoted as \textit{VehicleGAN+JML} or \textit{VehicleGAN$^*$+JML}. The results of Vehicle Re-ID on VeRi-776 and VehicleID datasets are shown in Table~\ref{tab:reid_776} and Table~\ref{tab:reid_id}, respectively. Our method gains better performance than the baseline and almost the best performance in all evaluation metrics.

\begin{table}[h]
\centering
\caption{Results of Vehicle Re-ID on VeRi-776~\cite{liu2016deep1} dataset. Higher mAP, Rank-1, and Rank-5 mean better performance. The best and second best performance are marked in {\color[HTML]{FE0000} \textbf{red}} and {\color[HTML]{3531FF} \textbf{blue}}.}
\begin{tabular}{c|cccc} 
\toprule
Methods  & mAP & Rank-1  & Rank-5 \\
\midrule
% EALN~\cite{lou2019embedding}  & 0.574 & 0.844 & 0.941  \\
% AAVER~\cite{khorramshahi2019dual} & 0.612 & 0.890 & 0.947  \\
RAM~\cite{liu2018ram}  & 0.615 & 0.886 & 0.940  \\
QD-DLF~\cite{zhu2019vehicle}  & 0.618 & 0.885 & 0.945  \\
VANet~\cite{chu2019vehicle} & 0.663 & 0.898 & 0.960  \\
BIR~\cite{wu2020background} & 0.704 & 0.905 & 0.971  \\
DFLNet~\cite{bai2020disentangled} & 0.732 & {\color[HTML]{3531FF} \textbf{0.932}}  &  {\color[HTML]{FE0000} \textbf{0.975}}   \\
\midrule
Baseline-ResNet50 & 0.703 & 0.916 & {\color[HTML]{3531FF} \textbf{0.973}}  \\
VehicleGAN*+JML & {\color[HTML]{3531FF} \textbf{0.736}}  & {\color[HTML]{FE0000} \textbf{0.936}} & {\color[HTML]{3531FF} \textbf{0.973}}   \\
VehicleGAN+JML  & {\color[HTML]{FE0000} \textbf{0.742}}  & {\color[HTML]{FE0000} \textbf{0.936}} & {\color[HTML]{3531FF} \textbf{0.973}}   \\
\bottomrule
\end{tabular}
\label{tab:reid_776}
\end{table}

\begin{table}[h]
\centering
\caption{Results of Vehicle Re-ID on VehicleID~\cite{liu2016deep2} dataset. Higher Rank-1, and Rank-5 mean better performance. The best and second best performance are marked in {\color[HTML]{FE0000} \textbf{red}} and {\color[HTML]{3531FF} \textbf{blue}}.}
\resizebox{1.0\columnwidth}{!}{ 
\begin{tabular}{c|cc|cc|ccc} 
\toprule
\multirow{2}{*}{Methods}  & \multicolumn{2}{c|}{Test800} & \multicolumn{2}{c|}{Test1600} & \multicolumn{2}{c}{Test2400} \\
         &  Rank-1  & Rank-5 & Rank-1  & Rank-5 & Rank-1  & Rank-5\\
\midrule
% EALN~\cite{lou2019embedding}& 0.751 & 0.881 & 0.718 & 0.839 & 0.693 & 0.814  \\
% AAVER~\cite{khorramshahi2019dual}& 0.747 & 0.938 & 0.686 & 0.900 & 0.635 & 0.856  \\
RAM~\cite{liu2018ram}& 0.752 & 0.915 & 0.723 & 0.870 & 0.677 & 0.845  \\
QD-DLF~\cite{zhu2019vehicle}& 0.723 & 0.925 & 0.707 & 0.889 & 0.641 & 0.834  \\
DF-CVTC~\cite{li2021attributes} & 0.752 & 0.881 & 0.722 & 0.844 & 0.705 & 0.821  \\
SAVER~\cite{khorramshahi2020devil}  & 0.799 & 0.952 & 0.776 & 0.911 & 0.753 & 0.883  \\
CFVMNet~\cite{sun2020cfvmnet}  & 0.814 & 0.941 & 0.773 & 0.904 & 0.747 & 0.887  \\
\midrule
Baseline-ResNet50 & 0.804 & 0.954 & 0.764 & 0.918 & 0.737 & 0.888  \\
VehicleGAN*+JML & {\color[HTML]{3531FF} \textbf{0.832}} & {\color[HTML]{FE0000} \textbf{0.966}} & 
 {\color[HTML]{FE0000} \textbf{0.783}} & {\color[HTML]{3531FF} \textbf{0.931}} & {\color[HTML]{FE0000} \textbf{0.759}} & {\color[HTML]{3531FF} \textbf{0.905}}  \\
VehicleGAN+JML & {\color[HTML]{FE0000} \textbf{0.835}} & {\color[HTML]{3531FF} \textbf{0.965}}  & {\color[HTML]{3531FF} \textbf{0.782}}  & {\color[HTML]{FE0000} \textbf{0.932}}  & {\color[HTML]{3531FF} \textbf{0.757}} & {\color[HTML]{FE0000} \textbf{0.906}}   \\
\bottomrule
\end{tabular}}
\label{tab:reid_id}
\end{table}

%------------------------------------------------------------------------- 
\section{Conclusions}\label{Sec:Conclusions}
This paper proposes a novel VehicleGAN for pose guided vehicle image synthesis, followed by a Joint Metric Learning framework to benefit vehicle Re-ID. The VehicleGAN utilizes a proposed AutoReconstruction as self-supervision for pose guided image synthesis. In this way, the proposed VehicleGAN is pair-flexible, working for either supervised (paired) or unsupervised (unpaired) setting. VehicleGAN is used to generate pose guided synthetic images with a unified target pose, which helps the feature-level fusion based Joint Metric Learning framework to learn vehicle perspective-invariant features, reducing the Re-ID recognition difficulties introduced by diverse poses of the same vehicles. Extensive experiments on two public datasets show that: 1) the proposed VehicleGAN can synthesize pose guided target image with high quality, 2) the proposed Joint Metric Learning framework obtains outstanding Re-ID accuracy with the assistance of VehicleGAN.

\ifCLASSOPTIONcaptionsoff
  \newpage
\fi

\small
\bibliographystyle{IEEEtran}
\bibliography{baolu.bib}

\end{document}